\documentclass{article}
\usepackage{ijcai26}

\usepackage{times}
\usepackage{soul}
\usepackage{url}
\usepackage[hidelinks]{hyperref}
\usepackage[utf8]{inputenc}
\usepackage[small]{caption}
\usepackage{graphicx}
\usepackage{amsmath}
\usepackage{amsthm}
\usepackage{booktabs}
\usepackage{algorithm}
\usepackage{algorithmic}
\usepackage[switch]{lineno}

\usepackage{xcolor}
\usepackage{subcaption}
\usepackage{enumitem}
\usepackage{amsfonts}

\usepackage{tabularx}
\usepackage{multirow}
\definecolor{highlight_red}{rgb}{1,0.4,0.5}
\definecolor{highlight_blue}{HTML}{74AED4}
\newcommand\tbred[1]{\textbf{\textcolor{highlight_red}{#1}}}
\newcommand\tblue[1]{\textcolor{highlight_blue}{#1}}

\usepackage{array}
\newcolumntype{Y}{>{\centering\arraybackslash}X}

\title{PDD-RRG: Posterior Diagnostic Decision for Study-level Radiology Report Generation}

\author{
Yang Yu$^{1}$\and
Yiming Ji$^{1}$\and
Bin Dai$^2$\and
Dong Zhang$^{1,3}$\footnotemark[1]\and
Zhiyong Zhou$^2$\and
Shoushan Li$^1$\and
Yakang Dai$^2$\\
\affiliations
$^1$School of Computer Science \& Technology, Soochow University\\
$^2$Suzhou Institute of Biomedical Engineering and Technology\\
$^3$Jiangsu Key Lab of Language Computing, Suzhou\\
\emails
dzhang@suda.edu.cn
}

\begin{document}

\maketitle
\renewcommand{\thefootnote}{\fnsymbol{footnote}}
\footnotetext[1]{Corresponding author.}
\renewcommand{\thefootnote}{\arabic{footnote}}

\begin{abstract}
    Automatic radiology report generation (RRG) aims to simulate the workflow of radiologists, assisting them in clinical diagnosis.
    However, existing methods often fall short in utilizing all information relevant to the examination, as is typically done in clinical practice.
    Although some works attempt to incorporate multi-view images and historical data, these additional inputs may sometimes lead to avoidable diagnostic errors on the contrary.
    To address these challenges, we introduce a decision-making stage after report generation for the first time and propose a Posterior Diagnostic Decision framework (PDD-RRG) to integrate potentially conflicting diagnoses.
    Specifically, we create various subsets of input data and utilize an existing RRG model to generate reports from different perspectives.
    Then the Bayesian posterior probability and the learned thresholds for each clinical observation are calculated to obtain an aggregated diagnostic conclusion, which is subsequently used to refine the generated report.
    Experiments on MIMIC-CXR demonstrate that our proposed PDD-RRG can effectively enhance the clinical efficacy of existing RRG models without any retraining.
\end{abstract}

\begin{figure}[t]
    \centering
    \setlength{\abovecaptionskip}{8pt}
    \setlength{\belowcaptionskip}{-18pt}
    \includegraphics[width=0.99\linewidth]{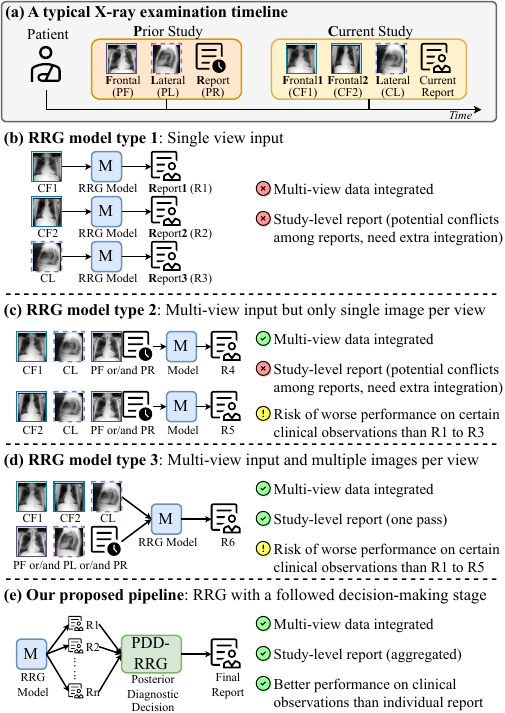}
    \caption{Pipelines of RRG models and our proposed framework.}
    \label{fig:intro}
\end{figure}

\section{Introduction}
Radiological examinations, led by chest X-rays, are widely used in clinical for the diagnosis of various diseases.
When drafting reports, radiologists must synthesize all information relevant to the current examination.
For the example in Figure \ref{fig:intro}(a), they have to read multi-view images and compare them with prior exams and reports (when available).
This process is both time-consuming and error-prone.
Consequently, automatic radiology report generation (RRG) which can provide high-quality draft reports, is attracting increasing attention.

RRG is intuitively an end-to-end text generation task.
The success of large language models (LLM) across various domains~\cite{DBLP:24cikm/lLiu,DBLP:25tmm/xcJu,DBLP:25acl-f/qwTang} has further reinforced this stereotype.
As Figure \ref{fig:intro}(b) shows, early methods~\cite{DBLP:18acl/CoAtt,DBLP:19cvpr/PPKED} typically relied on a single image, which diverges significantly from the real-world diagnostic workflow.
These generated drafts still require manual integration to resolve potential conflicts and supplement missing perspective information before serving as a study-level report.
Some recent studies~\cite{DBLP:23emnlp-f/SerraWD0O23,DBLP:25aaai/LLM-RG4} shown in Figure \ref{fig:intro}(c), have incorporated multi-view images and additional historical studies.
Nevertheless, due to the model limitation, they may still need to generate multiple reports to cover all information relevant to the examination.

\begin{figure}[t]
    \centering
    \setlength{\abovecaptionskip}{3pt}
    \setlength{\belowcaptionskip}{-10pt}
    \includegraphics[width=\linewidth]{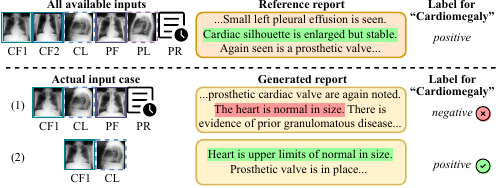}
    \caption{A real case on MAIRA-2 where additional historical information lead to an avoidable false negative for cardiomegaly.}
    \label{fig:case_maira-2}
\end{figure}

\begin{figure}[t]
    \centering
    \setlength{\abovecaptionskip}{5pt}
    \setlength{\belowcaptionskip}{-10pt}
    \includegraphics[width=\linewidth]{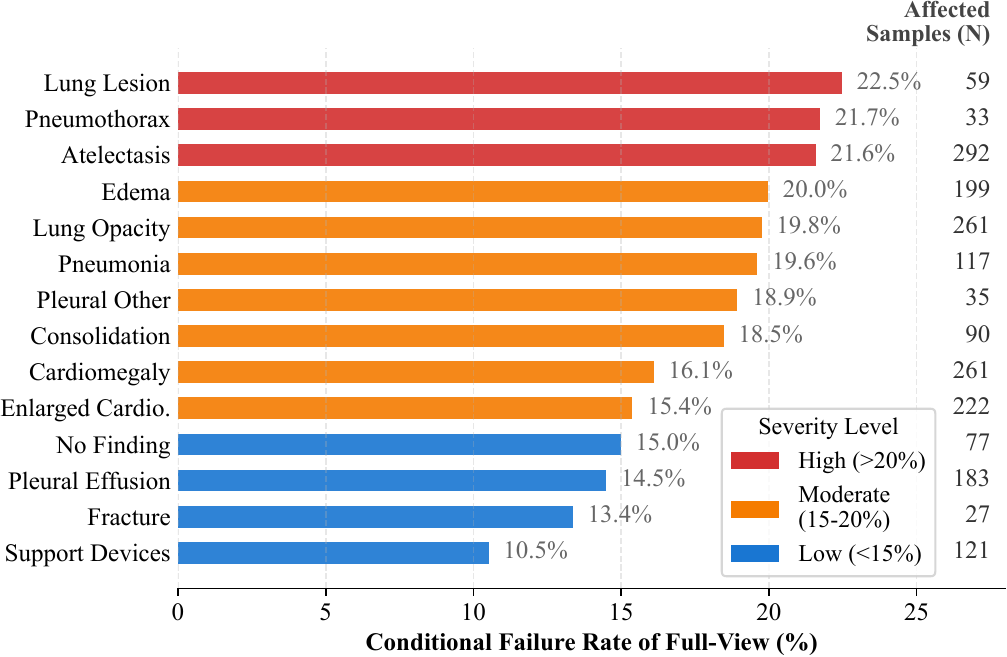}
    \caption{The proportion of diagnostic errors in various clinical observations that can be corrected with fewer inputs on MAIRA-2.}
    \label{fig:affected_ratio}
\end{figure}

Only a small number of approaches~\cite{nicolson2024longitudinal} are capable of ingesting all available cues in one pass to produce a full study-level report (as shown in Figure \ref{fig:intro}(d)).
However, this does not guarantee that their diagnoses are always optimal.
In fact, we observed that richer inputs can sometimes yield worse diagnostic results.
As the case illustrated in Figure \ref{fig:case_maira-2} on a Type-2 model MAIRA-2~\cite{DBLP:24corr/MAIRA-2}, introducing additional historical view causes a missed diagnosis of cardiomegaly that is correctly identified under simpler input.
We counted such cases in the MIMIC-CXR~\cite{DBLP:19corr/MIMIC-CXR} test set and found that they are far from rare.
As shown in Figure \ref{fig:affected_ratio}, for each of the 14 clinical observations extracted from reports, over 10\% of the samples suffer avoidable diagnostic errors when integrating all available inputs.

Superficially, this reflects a fusion flaw of current approaches, where additional cues may not be fully exploited and can even turn into noise that misleads the model.
However, at a deeper level, it reveals a persistent contradiction between the goal and reality of the RRG task.
As methods evole, we pack more cues into the input to attempt covering all available views in one pass, avoiding merging drafts generated from fragmented information.
But for certain findings, richer input can generate suboptimal results, so we may still need to consult the reports from simpler views to reach a sound diagnosis.
In short, to fully leverage a model’s diagnostic power, a report-merging step appears inevitable.

\begin{figure*}[t]
    \centering
    \setlength{\abovecaptionskip}{2pt}
    \setlength{\belowcaptionskip}{-11pt}
    \includegraphics[width=0.97\linewidth]{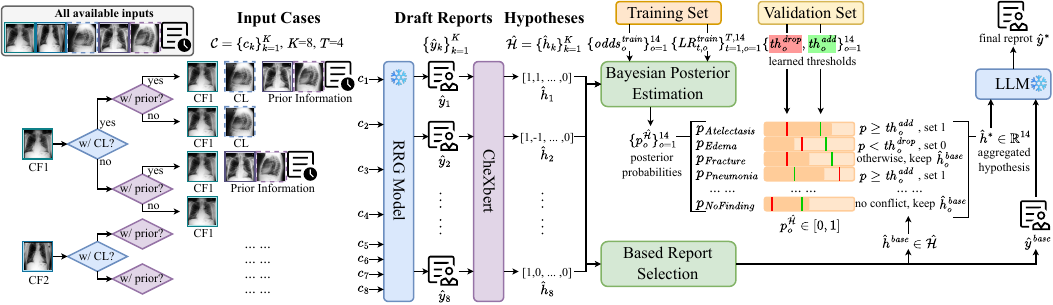}
    \caption{The architecture of our proposed pipeline. Note that $\hat{h}^*_{\mathit{No Finding}}$ is set to true only when all other revised observations are negative.}
    \label{fig:architecture}
\end{figure*}

Thus, we introduce this integration as an automated decision stage, implemented after report generation.
This shifts RRG from a single-step generation task to a pipeline of multi-path generations followed by a decision-level fusion, shown in figure \ref{fig:intro}(e).
On this basis, we propose PDD-RRG, a posterior diagnostic decision framework to aggregate conflicting diagnoses.
As Figure \ref{fig:architecture} shows, we extract 14 observation classes from reports, compute their Bayesian posteriors, and apply validation-tuned thresholds to finalize each diagnosis, guiding the new study-level report.
Experiments on three models demonstrate the effectiveness of our method.

Our contributions are stated as follows: 1) To the best of our knowledge, this is the first work to introduce a diagnostic decision layer into RRG, enabling the reconciliation of heterogeneous inputs and unlocking the latent diagnostic capability of backbone models. 2) We propose PDD-RRG, a likelihood-based aggregation framework that improves the utilization of clinically meaningful signals without requiring any model retraining. 3) We conduct extensive experiments and analyses on MIMIC-CXR, demonstrating the feasibility of posterior decision aggregation for multi-input fusion, bypassing the information fusion bottlenecks in LLM.
Code is available at \url{https://github.com/yynj98/PDD-RRG}.

\section{Related Works}
Radiology report generation (RRG) aims at clinically accurate reporting for examinations, yet while the target report is fixed, the input remains loosely defined.
Early and many recent studies~\cite{DBLP:19acl/CMAS,DBLP:23acl/ORGAN,DBLP:24cvpr/EKAGen,DBLP:25aaai/MPO} frame it as single-image captioning, ignoring the extra context reporting demands.
To mirror clinical routine, additional views such as lateral images~\cite{DBLP:21naacl/factENT,DBLP:22acl-f/CMM+RL,DBLP:conf/23bibm/LiXYCZFZG23,DBLP:25acl/NicolsonZDK25}, historical data~\cite{DBLP:24eccv/HERGen,DBLP:24mm/MSTF,DBLP:25aaai/HC-LLM,DBLP:25acl/RADAR}, and auxiliary text~\cite{DBLP:23nips-ml4h/Pragmatic_Llama,DBLP:24miccai/SEI} are progressively folded into the input.
MAIRA-2~\cite{DBLP:24corr/MAIRA-2} is the first to ingest all available views, yet concatenated features limit it to one image per view.
MLRG~\cite{DBLP:25cvpr/MLRG} anchors on one image and pools all other views via cross-attention, integrating every accessible cue.
Despite their success, no existing work noted that richer input can introduce avoidable errors.
To address this deficiency, our PDD-RRG aggregates reports from both lean and rich inputs, delivering a unified diagnosis that outperforms any single path.

\section{Methodology}
\subsection{Problem Formulation}

Given a study ${\mathcal{S}} = (\{x_m\}_{m=1}^M, z, \{x_n^{\mathit{pri}}\}_{n=1}^N,y^{\mathit{pri}},y)$, it consists of $M$ current images $x_m$, an auxiliary text $z$ such as “Indication”, $N$ previous images $x_n^{\mathit{pri}}$ with their report $y^{\mathit{pri}}$, and a reference report $y$. 
All images are frontal or lateral views, with $M$ typically ranging from 1 to 4.
Each study includes at least one current frontal image (CF), while $z$, $x^{\mathit{pri}}$, and $y^{\mathit{pri}}$ could all be absent.
The traditional RRG task is to learn a function $f_{R}(\cdot)$ that maps $(\{x_m\}_{m=1}^M, z, \{x_n^{\mathit{pri}}\}_{n=1}^N,y^{\mathit{pri}})$ or a portion of it to $y$ in one pass.

As shown in Figure \ref{fig:intro}(e), now we reframe the single-step RRG to a pipeline involving multi-path generations and a subsequent decision stage.
Given an RRG input $(\{x_m\}_{m=1}^M, z, \{x_n^{\mathit{pri}}\}_{n=1}^N,y^{\mathit{pri}})$, let $\mathcal{C}=\{c_k\}_{k=1}^K$ be a set of clinically plausible input cases derived from it, where each ${c_k}$ corresponds to a subset of all available inputs (e.g., a specific CF).
With a pre-trained RRG model $f_R(\cdot)$, each $c_k$ induces a draft $\hat{y}_k=f_R({c}_k)$ which can be regarded as a diagnostic hypothesis.
These hypotheses may differ in terms of identified findings, confidence, or completeness. Rather than assuming that any single hypothesis $\hat{y}_k$ is optimal, we aim to infer a final diagnostic decision.
Formally,
\begin{equation}
    \hat{y}^* = f_D \left( \{\hat{y}_k\}_{k=1}^{K} \right)
\end{equation}
where $f_{D}$ is a posterior decision function that aggregates multiple hypotheses into a unified and robust diagnostic outcome.

\subsection{Multi-input Diagnostic Hypothesis Generation}\label{sec:input_gen}

To obtain diverse diagnostic evidence, we need to first generate multiple input cases, where each $c_k \in \mathcal{C}$ reflects a realistic diagnostic scenario.
To achieve it, we pair each CF with a lateral image and/or prior information, deriving up to four input configurations as shown in Figure \ref{fig:architecture} (auxiliary text $z$ is always used if available).
All four types exist in real datasets and can be directly received by Type-2 or Type-3 models in Figure \ref{fig:intro}.
We then generate concrete cases according to input types, so that each $c_k$ has its own type, too.
This is formulated as $\mathit{type}(c_k) \in \{1,2,\cdots,T\}$, where $T$ denotes the number of input types.
Apart from the above four, 
we can also try different parts of prior reports (e.g., full vs. only “FINDINGS”) to obtain eight or even more input types.

For each ${c}_k$ $\in$ $\mathcal{C}$, the adopted RRG model generates a draft report $\hat{y}_k$.
To obtain a structured diagnosis, we utilize CheXbert~\cite{DBLP:20emnlp/CheXbert} to extract the 14-category labels $\hat{h}_k \in \mathbb{R}^{14}$ from $\hat{y}_k$ and take it as the corresponding hypothesis.
Each label $\hat{h}_{k,o} \in \{1,0,-1,\mathit{null}\}$ in $\hat{h}_k$ encodes the status (among positive, negative, uncertain, and not mentioned) of a specific clinical observation listed in Figure \ref{fig:affected_ratio} (e.g., Edema).
This process results in a set of diagnostic vectors $\hat{\mathcal{H}}=\{\hat{h}_k\}_{k=1}^K$.
Notably, all these hypotheses are generated independently based on the same model, without any retraining or architectural modification, allowing PDD-RRG to be applied as a lightweight post-hoc decision module.

This step can be interpreted as eliciting multiple latent reasoning paths of the same model under different informational conditions. Due to varying visibility of pathological cues and model attention biases, these hypotheses may emphasize different findings, exhibit different levels of confidence, or even conflict with each other. Such diversity, instead of being discarded, serves as the foundation for posterior diagnostic decision-making.

\subsection{Posterior Diagnostic Decision Framework}
In our proposed PDD-RRG framework, we aggregate all hypotheses $\hat{h}_k \in \hat{\mathcal{H}}$ to obtain a revised diagnosis $\hat{h}^*$, and synthesize a revised report $\hat{y}^*$ based on it.
Specifically, we first select a based report $\hat{y}^{\mathit{base}}$ as the revision template, and calculate the Bayesian posterior $p^{\hat{\mathcal{H}}}_o$ for each observation $o$.
If there is uncertain or conflicting (1 and 0/$\mathit{null}$) predictions for $o$, 
we determine the final diagnosis $\hat{h}^*_o$ (positive or negative) based on its corresponding $p^{\hat{\mathcal{H}}}_o$ and two thresholds ($\mathit{th}_o^{\mathit{drop}}$, $\mathit{th}_o^{\mathit{add}}$) learned on the validation set.
Finally, we employ a LLM to revise $\hat{y}^{\mathit{base}}$ according to $\hat{h}^*$, resulting in the final report $\hat{y}^*$.

Rather than assuming equal contributions from all hypotheses like majority voting, PDD-RRG explicitly accounts for the observation-specific diagnostic reliability of each input case, thus inferring a final output that is more stable and clinically reliable than any individual hypothesis.
In this section, we will elaborate on the above points in order.

\subsubsection{Based Report Selection via Consensus Maximization}
For each study, multiple reports are derived from varying input.
Instead of relying on the report from the most complete input, we select $\hat{y}^{\mathit{base}}$ as a consensus report that best reflects the model's overall judgment.

We evaluate consensus through the similarity between the diagnosis vectors $\hat{h}_k \in \hat{\mathcal{H}}$.
For each label $\hat{h}_{k,o} \in \{1,0,-1,\mathit{null}\}$ in $\hat{h}_k$, we define a pairwise label-consistency function $\mathit{sim}(\hat{h}_{i,o}, \hat{h}_{j,o}) \in [0,1]$ to reward similar or identical findings on the observation $o$:
\begin{equation}
    \mathit{sim}(\hat{h}_{i,o}, \hat{h}_{j,o}) = \begin{cases}
        1, &\hat{h}_{i,o} = \hat{h}_{j,o} = 0/1;\\
        0.5, &\hat{h}_{i/j,o} =-1 \wedge \hat{h}_{j/i,o} = 0/1;\\
        0, &\text{otherwise}.
    \end{cases}
\end{equation}
Then the $\mathit{sim}(\hat{h}_{i,o}, \hat{h}_{j,o})$ is averaged across all mentioned observations (aka., $\hat{h}_{k,o} \neq \mathit{null}$).
Denoting $N_{i,j}^{\mathit{null}}$ as the number of labels unmentioned by either side, $\mathit{sim}(\hat{h}_{i}, \hat{h}_{j})$ is computed as:
\begin{equation}
    \mathit{sim}(\hat{h}_{i}, \hat{h}_{j}) = \frac{1}{14- N_{i,j}^{\mathit{null}}} \sum_{o=1}^{14}{\mathit{sim}(\hat{h}_{i,o}, \hat{h}_{j,o})}
\end{equation}
Given a candidate $\hat{h}_k$, we sum up all its similarities between available reports to measure consensus:
\begin{equation}
    \mathit{score}(\hat{h}_k) = \sum_{i=1,i\ne k}^{K} \mathit{sim}(\hat{h}_k,\hat{h}_i)
\end{equation}
and select the report that maximizes the score as the $\hat{y}^{\mathit{base}}$.
\begin{equation}
    \hat{h}^{\mathit{base}} = \mathop{\arg\max}_{\hat{h}_k \in \mathcal{H}} \ \mathit{score}(\hat{h}_k)
\end{equation}
When multiple candidates tie, the shorter report is chosen to reduce verbosity and potential hallucinations.
This strategy yields a report that best aligns with the majority of the model’s own predictions, providing a stable anchor for posterior refinement.

\subsubsection{LR-based Bayesian Posterior Probability Estimation}

For each clinical observation in the 14 categories, we estimate the diagnostic reliability of each input type using the training data.
We collect hypotheses derived from studies in the training set, constructing a large prediction set $\hat{\mathcal{H}}_{\mathit{train}} = \{\hat{h}_k\}_{k=1}^{K_\mathit{train}}$.
Given the large number of samples, we build only one input case per current frontal (CF) image that integrates the maximum available views among $T$ input types.
After the based report selection, we treat the label “not mentioned” as a negative finding, so all $\hat{h}_{k,o} \in \{1,0,-1\}$.
The ground-truth labels $h_k \in \mathbb{R}^{14}$ for each hypothesis likewise form the $\mathcal{H}_{\mathit{train}} = \{h_k\}_{k=1}^{K_\mathit{train}}$, where $h_{k,o} \in \{1,0\}$.

For each input type $t\in \{1,2,\dots,T\}$ and observation $o\in\{1,2,\dots,14\}$, we compute the prior recall $R$ and false positive rate $\mathit{FP}$ over the training set as:
\[
R^{\mathit{train}}_{t,o} = P(\hat{h}_{k,o}=1 \: |\: h_{k,o}=1, \mathit{type}(c_k)=t, h_k\in\mathcal{H}_\mathit{train})
\]
\[
\mathit{FP}^{\mathit{train}}_{t,o} = P(\hat{h}_{k,o}=1 \: |\: h_{k,o}=0, \mathit{type}(c_k)=t, h_k\in\mathcal{H}_\mathit{train})
\]
These statistics quantify how strongly a positive prediction from a given input type $t$ supports the presence of an observation $o$. 
We then separately define the likelihood ratio $\mathit{LR}$ for positive and negative observations.
\begin{equation}
    \mathit{LR}_{t,o}^{\mathit{train}+} = \frac{R^{\mathit{train}}_{t,o}}{\mathit{FP}^{\mathit{train}}_{t,o}},\ \mathit{LR}_{t,o}^{\mathit{train}-} = \frac{1-R^{\mathit{train}}_{t,o}}{1-\mathit{FP}^{\mathit{train}}_{t,o}}
\end{equation}

We adopt $p^{\mathit{train}}_o$ to represent the prior positive probability $P(y=1)$ for each observation $o$:
\begin{equation}
    p^{\mathit{train}}_o = P(h_{k,o}=1 \mid h_k\in\mathcal{H}_\mathit{train})
\end{equation}
and obtain the corresponding prior odds as
\begin{equation}
    \mathit{odds}^{\mathit{train}}_o = \frac{p^{\mathit{train}}_o}{1-p^{\mathit{train}}_o}
\end{equation}

Finally, we estimate the posterior log-odds for the given hypotheses $\hat{\mathcal{H}} = \{\hat{h}_k\}^K_{k=1}$ as
\begin{equation}\label{eq:1}
    \log{\mathit{odds}^{\hat{\mathcal{H}}}_o} = \log{\mathit{odds}^{\mathit{train}}_o} + \sum_{\hat{h}_{k}\in \hat{\mathcal{H}}}{\alpha \log{\mathit{LR}_{t,o}^{\mathit{train}}}}
\end{equation}
and achieve the posterior positive probability of each observation based on the definition of odds.
\begin{equation}
    p^{\hat{\mathcal{H}}}_o = P(\hat{h}_{k,o}=1 \mid \hat{h}_k\in \hat{\mathcal{H}}) = \frac{1}{1+\exp{(-\log{\mathit{odds}^{\hat{\mathcal{H}}}_o})}}
\end{equation}

Note that in Equation \ref{eq:1}, both the values of $\mathit{LR}_{t,o}^{\mathit{train}}$ and $\alpha$ depend on the current prediction $\hat{h}_{k,o}$. We select $\mathit{LR}_{t,o}^{\mathit{train}}$ via
\begin{equation}
    \mathit{LR}_{t,o}^{\mathit{train}} = \begin{cases}
        \mathit{LR}_{t,o}^{\mathit{train}+}, &\hat{h}_{k,o}\in \{1,-1\} \wedge \mathit{type}(c_k)=t; \\
        \mathit{LR}_{t,o}^{\mathit{train}-}, &\hat{h}_{k,o} = 0 \wedge \mathit{type}(c_k)=t.
    \end{cases}
\end{equation}
and set $\alpha$ to 1 or 1.2 when $\hat{h}_{k,o}$ equals 1 or 0, where the mild amplification factor is to prevent overconfidence arising from sporadic false positives. 
For uncertain predictions, we assign the condition-specific $\alpha$ with an empirical probability as:
\begin{equation}
	\alpha_{t,o} = \frac{P(\hat{h}_{k,o}= -1 \: |\: h_{k,o}=1, \mathit{type}(c_k)=t, h_k\in\mathcal{H}_\mathit{train})}{P(\hat{h}_{k,o}= -1 , \mathit{type}(c_k)=t, \hat{h}_k\in \hat{\mathcal{H}}_\mathit{train})}
\end{equation}
This effectively softens the contribution of uncertain evidence, allowing it to influence posterior inference while preventing unstable amplification.

\subsubsection{Diagnostic Decision and Threshold Calibration}
Given posterior probabilities $\{p^{\hat{\mathcal{H}}}_o\}^{14}_{o=1}$, PDD-RRG determines the final status (positive or negative) for each observation based on learned thresholds.
Considering that different diseases exhibit highly heterogeneous prevalence, uncertainty, and hallucination patterns, we therefore learn two observation-specific thresholds rather than a global one: 1) a drop threshold $\mathit{th}_o^{\mathit{drop}}$ to suppress unreliable positives, and 2) an add threshold $\mathit{th}_o^{\mathit{add}}$ to introduce missing positives.
In this way, PDD-RRG adapts to each disease’s posterior uncertainty and enables asymmetric control of false positives (FP) and false negatives (FN), which is critical in clinical practice.

Assuming that we already have the learned thresholds, the decision for each observation $o$ is defined as:
\begin{equation}
    \hat{h}_o^* = \begin{cases}
        0, &p^{\hat{\mathcal{H}}}_o < \mathit{th}_o^{\mathit{drop}};\\
        1, &p^{\hat{\mathcal{H}}}_o \ge \mathit{th}_o^{\mathit{add}};\\
        \hat{h}^{\mathit{base}}_{o}, &\text{otherwise}.
    \end{cases}
\end{equation}
If both thresholds are met ($\mathit{th}_o^{\mathit{add}} \le p^{\hat{\mathcal{H}}}_o < \mathit{th}_o^{\mathit{drop}}$), we set $\hat{h}_o^*$ to 1, since missed diagnoses carry heavier clinical penalties.
If neither threshold is met ($\mathit{th}_o^{\mathit{drop}} \le p^{\hat{\mathcal{H}}}_o < \mathit{th}_o^{\mathit{add}}$), we default to the judgment in $\hat{y}^{\mathit{base}}$.
To calibrate them, we apply our PDD-RRG to the validation set, setting each $\mathit{th}_o^{\mathit{drop/add}}$ from 0 to 1 in steps of 0.05, and select the value that maximizes the $F_1$-score over all hypotheses $\{\hat{h}^*_o\}_{\mathit{val}}$  as the final threshold:
\begin{equation}
    \mathit{th}_o^{\mathit{drop/add}} = \mathop{\arg\max}_{\mathit{th}_o^{\mathit{drop/add}} \in [0,1]} \ F_1 \left(\{\hat{h}^*_o\}_{\mathit{val}},\{h_o\}_{\mathit{val}}\right)
\end{equation}

\subsubsection{Report Refinement Based on Revised Diagnosis}
Having the final diagnosis $\hat{h}^*$ for 14 observations, we simply leverage a frozen online LLM (e.g., GPT-4) to revise the based report $\hat{y}^{\mathit{base}}$, achieving the refined report $\hat{y}^{*}$.
\begin{equation}
    \hat{y}^{*} = \text{LLM}(\hat{h}^*,\hat{y}^{\mathit{base}})
\end{equation}

\begin{table*}[t]
\centering
\footnotesize
\scriptsize
\setlength{\abovecaptionskip}{5pt}
\setlength{\belowcaptionskip}{-10pt}
\renewcommand{\arraystretch}{0.98}
\begin{tabular}{llrrrrrrrr}
\toprule
Model & Method & BLEU-1 & BLEU-4 & METEOR & ROUGE-L & Mac-F1$_{14}$ & Mic-F1$_{14}$ & Mac-F1$_5$ & Mic-F1$_5$ \\
\midrule

\multirow{4}{*}{MAIRA-2}
& Raw      & \tbred{33.89} & \tbred{13.19} & \tbred{34.30} & \tbred{31.25} & 39.34 & 56.09 & 46.84 & 56.68 \\
& Selected & 28.35 & 10.05 & 29.52 & 27.95 & 39.34 & 57.07 & 47.65 & 58.24 \\
& PDD      & $\text{--}$ & $\text{--}$ & $\text{--}$ & $\text{--}$ & \tbred{42.94} & \tbred{59.85} & \tbred{51.18} & \tbred{60.03} \\
& PDD-report & \tblue{31.20} & \tblue{10.52} & \tblue{31.50} & \tblue{28.03} & \tblue{42.25} & \tblue{59.22} & \tblue{50.82} & \tblue{59.85} \\
\midrule

\multirow{4}{*}{LLM-RG4}
& Raw      & \tbred{34.92} & \tbred{12.42} & \tbred{33.85} & \tbred{30.24} & 41.67 & 58.70 & 52.29 & 59.96 \\
& Selected & 31.01 & 10.29 & 30.76 & \tblue{28.96} & 41.72 & 58.65 & 51.99 & 60.49 \\
& PDD      & -- & -- & -- & -- & \tbred{42.56} & \tblue{59.09} & \tbred{53.13} & \tbred{61.44} \\
& PDD-report & \tblue{32.69} & \tblue{10.63} & \tblue{32.16} & 28.76 & \tblue{42.08} & \tbred{59.66} & \tblue{52.95} & \tbred{61.44} \\
\midrule

\multirow{4}{*}{MLRG}
& Raw      & \tblue{34.90} & \tbred{12.14} & \tblue{33.16} & \tbred{29.62} & 32.65 & 52.80 & 44.54 & 53.95 \\
& Selected & 34.17 & 11.86 & 32.63 & \tblue{29.55} & 32.93 & 53.00 & 45.30 & 54.82 \\
& PDD      & -- & -- & -- & -- & \tbred{35.37} & \tblue{53.69} & \tbred{46.58} & \tbred{55.78} \\
& PDD-report & \tbred{35.13} & \tblue{11.91} & \tbred{33.63} & 29.33 & \tblue{35.32} & \tbred{54.50} & \tblue{46.47} & \tblue{55.72} \\
\bottomrule
\end{tabular}
\caption{Main performance comparison on the MIMIC-CXR test set.
Following common RRG evaluation practice, Atelectasis, Cardiomegaly, Consolidation, Edema, and Pleural Effusion are used to compute F1$_5$. 
Best and second best results are highlighted in \tbred{red} and \tblue{blue} respectively.
}
\label{tab:main_results_compact}
\end{table*}

Looking back at our proposed PDD-RRG, we explicitly decouple diagnostic decision-making from report generation, mitigating the instability introduced by input selection and generation bias. 
As a result, the final diagnosis reflects a more comprehensive utilization of available medical evidence and provides more robust support for clinical judgment.

\section{Experimentation}

\subsection{Experimental Setup}\label{sec:Exp_setip}
\textbf{Dataset.}
MIMIC-CXR~\cite{DBLP:19corr/MIMIC-CXR} is a widely used RRG dataset comprising multi-view images and longitudinal patient information.
Following previous work, we treat the section “FINDINGS” in reports as our target, conducting all experiments on the filtered official split (resulting in 152,173/1,196/2,347 studies in the train/val/test set).

\textbf{Base RRG Models.}
To better evaluate generality, we apply PDD-RRG to three state-of-the-art baselines: MAIRA-2~\cite{DBLP:24corr/MAIRA-2}, LLM-RG4~\cite{DBLP:25aaai/LLM-RG4}, and MLRG~\cite{DBLP:25cvpr/MLRG}.
MAIRA-2 and LLM-RG4 are Type-2 models shown in Figure \ref{fig:intro}, we adopt the eight and four input types introduced in Section \ref{sec:input_gen}, respectively.
For Type-3 MLRG, we similarly create four input types by pairing each anchor image with optional auxiliary images and/or prior information.
Note that PDD-RRG is applied strictly as a post-hoc diagnostic decision module, meaning that all base models remain frozen and no retraining or architectural modification is required.

\textbf{Evaluation Metrics.} 
We evaluate performance on both NLG and Clinical Efficacy (CE) metrics.
We report BLEU-1, BLEU-4, METEOR, and ROUGE-L, while focusing primarily on the CE scores~\cite{DBLP:20emnlp/R2Gen} derived by CheXbert.
Specifically, we map “uncertain” labels to negative and compute macro- and micro-average F1 scores for 14 observations extracted from reports.

\subsection{Main Results on Various Base Models}\label{sec:main_results}
We apply PDD-RRG to three representative RRG models: MAIRA-2, LLM-RG4, and MLRG. For each model, we compare four scenarios under multi-input settings:
(i) the original output by richest input (Raw), 
(ii) the most self-consistent report $\hat{y}^{\mathit{base}}$ selected in our approach (Selected), 
(iii) the diagnostic labels $\hat{h}^*$ produced by our posterior decision module (PDD), and 
(iv) the reconstructed report $\hat{y}^*$ generated from $\hat{h}^*$ (PDD-report).
Table \ref{tab:main_results_compact} shows the results of these scenarios across all evaluation metrics.
Two key observations emerge.

\emph{First, PDD-RRG consistently improves clinical performance across all base models}. These uniform gains, achievable regardless of the underlying architecture, demonstrate the strong model-agnostic generalization of our approach.

\emph{Second, Selected frequently outperforms Raw on clinical metrics}. This confirms that the most information-complete input does not necessarily yield the most reliable diagnosis, validating the necessity of explicit posterior decision-making under heterogeneous inputs.

Regarding language quality, PDD-report shows slightly lower NLG scores than Raw but consistently outperforms Selected, which serves as the backbone for report reconstruction.
This indicates that label-aligned report reconstruction preserves, and in some cases improves, textual quality. Overall, \emph{PDD-RRG provides robust, architecture-independent improvements in clinical performance while maintaining competitive language performance}, validating its suitability as a general plug-and-play diagnostic refinement layer for RRG.

\begin{table}[t]
\centering
\footnotesize
\scriptsize
\setlength{\abovecaptionskip}{5pt}
\setlength{\belowcaptionskip}{-10pt}
\renewcommand{\arraystretch}{0.98}
\begin{tabular}{lrrrr}
\toprule
Method & Mac-F1$_{14}$ & Mic-F1$_{14}$ & Mac-F1$_5$ & Mic-F1$_5$ \\
\midrule
Baseline (Selected)    & 39.34 & 57.07 & 47.65 & 58.24 \\
PDD                    & 42.94 & \textbf{59.85} & \textbf{51.18} & \textbf{60.03} \\
PDD w/o LR             & 42.59 & 59.17 & 50.90 & 59.32 \\
PDD w/o OSDT  & \textbf{43.48} & 59.15 & 51.16 & 59.55 \\
\bottomrule
\end{tabular}
\caption{Ablation study of different components on MAIRA-2.}
\label{tab:ablation}
\end{table}

\subsection{Ablation Study}
We conduct ablation studies to evaluate two key components of PDD-RRG: Likelihood-Ratio (LR)-based posterior modeling and Observation-Specific Decision Thresholds (OSDT). Results on MAIRA-2 are reported in Table \ref{tab:ablation}.

\textbf{Effect of Likelihood-Ratio Modeling.} 
We replace the LR-based formulation with a frequency-based aggregation scheme (“w/o LR”), computing disease confidence as the frequency of non-negative predictions in $\hat{\mathcal{H}}$ (uncertain as 0.5).
Although frequency aggregation improves over the selected report, it fails to match LR modeling. These results confirm that \emph{PDD-RRG’s gains stem from explicitly modeling heterogeneous diagnostic evidence rather than naive aggregation}.

\textbf{Effect of Observation-Specific Decision Thresholds.}
We further compare PDD with a variant that applies two global thresholds $\mathit{th}^{\mathit{drop/add}}$ to all observations (“w/o OSDT”). While the unified-threshold variant yields a slightly higher macro F1$_{14}$, \emph{the gain is unstable and likely coincidental, failing to generalize to other clinically relevant metrics}. This demonstrates that \emph{a global decision boundary is insufficient to capture the substantial inter-disease variability}.

\begin{table}[t]
\centering
\footnotesize
\scriptsize
\setlength{\abovecaptionskip}{3pt}
\setlength{\belowcaptionskip}{-8pt}
\renewcommand{\arraystretch}{0.98}

\begin{tabular}{llrrrr}
\toprule
Model & Method & Mac-F1$_{14}$ & Mic-F1$_{14}$ & Mac-F1$_5$ & Mic-F1$_5$ \\
\midrule

\multirow{2}{*}{MAIRA-2}
& Vote & 38.98 & 57.89 & 47.13 & 58.24 \\
& PDD  & \tbred{42.94} & \tbred{59.85}
       & \tbred{51.18} & \tbred{60.03} \\
\midrule

\multirow{2}{*}{LLM-RG4}
& Vote & 41.30 & 58.81 & 51.18 & 60.41 \\
& PDD  & \tbred{42.56} & \tbred{59.09} 
       & \tbred{53.13} & \tbred{61.44} \\
\midrule

\multirow{2}{*}{MLRG}
& Vote & 32.64 & 53.14 & 45.23 & 54.60 \\
& PDD  & \tbred{35.37} & \tbred{53.69}
       & \tbred{46.58} & \tbred{55.78} \\
\bottomrule
\end{tabular}
\caption{Comparison between Voting and PDD-RRG across three backbones.
Best results for each backbone are highlighted in \tbred{red}.}
\label{tab:vote_vs_ours}
\end{table}

\subsection{Comparison with Majority Voting}
Conventional RRG is typically formulated as an end-to-end text generation problem without an explicit diagnostic decision layer, and thus prior work has not addressed how to reconcile potentially conflicting outputs arising from multiple inputs. For a principled comparison, we adopt majority voting as a representative and widely used aggregation baseline.

\emph{As shown in Table \ref{tab:vote_vs_ours}, PDD consistently outperforms Voting across all base models}. This demonstrates that \emph{equal-weight voting fails to effectively aggregate heterogeneous diagnostic evidence}, especially in clinical scenarios where different findings are preferentially revealed by different input configurations. In such cases, Voting suffers from a \emph{vote dilution} effect, where highly informative but infrequent signals are overwhelmed by numerous weak or uninformative votes.

Overall, these results confirm that majority voting is suboptimal in medical decision aggregation, whereas \emph{PDD-RRG provides a more reliable and clinically aligned alternative for multi-input radiology report generation}.

\subsection{Decision Behavior Under Conflicting Inputs}\label{sec:analysis}
While Section \ref{sec:main_results} demonstrates the effectiveness of PDD-RRG, this section investigates the underlying decision mechanisms. We analyze how different aggregation strategies resolve conflicting hypotheses, revealing a fundamental asymmetry in distinct error types handing. We compare the selected base output with two post-hoc strategies on MAIRA-2:

\textbf{Silence-biased Perception (Base).} 
The selected report $\hat{y}^{\mathit{base}}$ inherits the model's training bias. It defaults to silence under uncertainty, leading to high precision but low recall.

\textbf{Consensus-driven Conservatism (Vote).} 
Majority voting aggregates views based on agreement. While effective at suppressing noise, it indiscriminately penalizes minority but potentially correct signals, reinforcing collective silence.

\textbf{Evidence-weighted Decision (PDD).} 
In contrast, PDD-RRG re-weights alternative views via likelihood ratios, allowing reliable minority evidence to override defaults, shifting the decision boundary from silence toward detection.

\subsubsection{Conflict Definition and Recoverability}
We formalize decision behavior by comparing the selected $\hat{h}^{\mathit{base}}$ against other hypotheses in $\hat{\mathcal{H}}$.
A \emph{diagnostic conflict} occurs when their predictions disagree. 

Errors in $\hat{h}^{\mathit{base}}$ are categorized as false negatives (FN)
, where GT is positive but predicted negative, and false positives (FP), where GT is negative but predicted positive.
An error is deemed recoverable if at least one $\hat{h}_k \in \hat{\mathcal{H}}$ predicts correctly, defining the theoretical upper bound achievable by post-hoc decision rules in our multi-view setting.

\subsubsection{Asymmetric Error Recoverability in Multi-view RRG}
We first establish the theoretical upper bound for post-hoc correction. As shown in Table \ref{tab:recoverability}, all based reports contain 2,315 FNs and 2,019 FPs, of which 37.88\% and 67.41\% are recoverable, respectively. This indicates that hallucinated findings are more easily contradicted across views, while missed diagnoses require explicit positive evidence. Consequently, FNs are intrinsically harder to recover than FPs.

\begin{table}[t]
\centering
\footnotesize
\scriptsize
\setlength{\abovecaptionskip}{3pt}
\setlength{\belowcaptionskip}{-8pt}
\renewcommand{\arraystretch}{0.98}
\begin{tabular}{lrrr}
\toprule
Error Type & Total & Recoverable & Rate (\%) \\
\midrule
False Negative (FN) & 2315 & 877  & 37.88 \\
False Positive (FP) & 2019 & 1361 & 67.41 \\
\bottomrule
\end{tabular}
\caption{Theoretical recoverability of errors in the test set.}
\label{tab:recoverability}
\end{table}

\begin{table}[t]
\centering
\footnotesize
\scriptsize
\setlength{\abovecaptionskip}{3pt}
\setlength{\belowcaptionskip}{-8pt}
\renewcommand{\arraystretch}{0.98}
\begin{tabular}{lrrrr}
\toprule
Repair Type & Vote & PDD & Vote-only & PDD-only \\
\midrule
FN recovery & 155 & \textbf{538} & \tbred{4} & \tbred{387} \\
FP recovery & \textbf{542} & 203 & 384 & 45 \\
\bottomrule
\end{tabular}
\caption{Comparison of FN and FP recovery under Voting and PDD-RRG. “Vote-only” and “PDD-only” denote cases uniquely repaired by each method.}
\label{tab:vote_pdd_recovery}
\end{table}

Within this recoverable pool, Table \ref{tab:vote_pdd_recovery} reveals a fundamental asymmetry in how Voting and PDD repair errors:

\textbf{PDD Strictly Dominates in FN Recovery.}
Among recoverable FNs, PDD recovers 61.35\% of cases, compared to only 17.67\% achieved by Voting. More importantly, \emph{PDD uniquely recovers 387 cases, whereas Voting contributes virtually no unique value (4 cases)}.

This strict dominance highlights a limitation of consensus-based decision rules in high-uncertainty medical settings. FN errors typically arise from weak or view-dependent evidence, where only a minority of input configurations reveal the abnormality. Voting indiscriminately suppresses minority signals, while PDD-RRG's likelihood-ratio weighting exploits weak but reliable evidence to rescue missed diagnoses.

\textbf{Voting only Looks Better in FP Suppression.}
Although Voting suppresses a larger fraction of FPs than PDD (39.82\% vs. 14.92\%), but \emph{this apparent advantage is largely an artifact of protocol-driven silence mapping and passive consensus bias, rather than explicit diagnostic refutation}.

First, radiology data are highly imbalanced toward negative labels and silence. Trained on such data, models tend to default to silence when visual evidence is weak or ambiguous. Voting amplifies this bias by interpreting cross-view omission as negative evidence. Since the test set follows the same negative-dominant distribution, Voting gains a natural albeit passive advantage by simply aligning with the majority class.

Second, many “suppressed” FPs are not truly contradicted by other views. In most alternative $\hat{h}_k$, the model omits the finding rather than negating it. Standard evaluation protocols conventionally map unmentioned categories to negative labels. Consequently, when the majority of views are silent, they form a “consensus of omission”. If the ground truth is also unmentioned, this is counted as a correct suppression. However, this represents a \emph{coincidental agreement} driven by default mapping rules, not a confirmed diagnostic exclusion.

Third, hallucinated positives are typically \emph{stochastic and unstable across views}, making them particularly vulnerable to consensus-based dilution. 

In contrast, PDD-RRG enables targeted recovery of clinically critical abnormalities that are weakly expressed and thus silenced by the majority. Notably, approximately 20\% of the FPs repaired by PDD are unique cases that Voting fails to recover, confirming that PDD-RRG provides essential corrective power where conservative consensus fails.

\subsubsection{Clinical Trade-Offs Under Asymmetric Error Correction}
\begin{figure}[t]
    \centering
    \setlength{\abovecaptionskip}{5pt}
    \setlength{\belowcaptionskip}{-8pt}
    \includegraphics[width=1.0\linewidth]{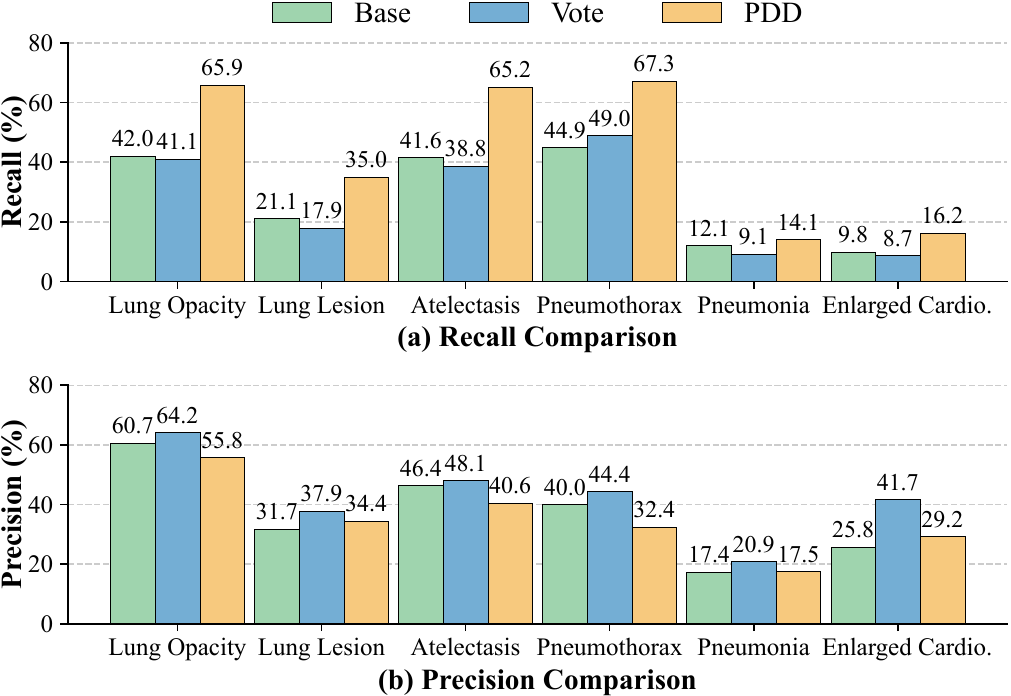}
    \caption{Recall (a) and Precision (b) comparisons. The numbers above bars indicate the corresponding post-decision values.}
    \label{fig:Metrics_evidence}
\end{figure}

Figure \ref{fig:Metrics_evidence} presents per-disease Recall and Precision for the selected $\hat{h}^{\mathit{base}}$, Voting, and PDD under multi-view inputs. Due to space constraints, we show six representative diseases here (full results are provided in our code repository).

As visualized in Figure \ref{fig:Metrics_evidence}, both Base and Voting strategy exhibit a “false-good” phenomenon: high Precision but very low Recall (e.g., $<$25\% for Pneumonia and Lung Lesion). 
In stark contrast, \emph{PDD achieves a systematic recall surge across all diseases}. For instance, Recall for Lung Opacity and Pneumothorax jumps by over 20 percentage points (e.g., Pneumothorax: 44.9\% $\rightarrow$ 67.3\%). This demonstrates that PDD-RRG effectively activates valid positive signals that were suppressed by the consensus mechanism. These gains are most pronounced for \emph{subtle or focal abnormalities} such as Lung Opacity, Atelectasis and Lung Lesion. 


Figure \ref{fig:Metrics_evidence}(b) shows a moderate Precision drop for PDD, expected due to the denominator effect: Base and Voting predict few positives (small denominator), resulting in inflated precision. By activating potential findings, PDD-RRG expands the denominator. \emph{From a clinical perspective, this trade-off is highly desirable: missing a life-threatening condition is far costlier than raising a false alarm}.

Overall, PDD-RRG transforms the model \emph{from a passive silence-biased predictor into a clinically viable screening system}, trading a marginal precision drop for a critical breakthrough in sensitivity.

\subsubsection{Lesion-level Repair Analysis}

\begin{figure}[t]
    \centering
    \setlength{\abovecaptionskip}{5pt}
    \setlength{\belowcaptionskip}{-8pt}
    \includegraphics[width=1.0\linewidth]{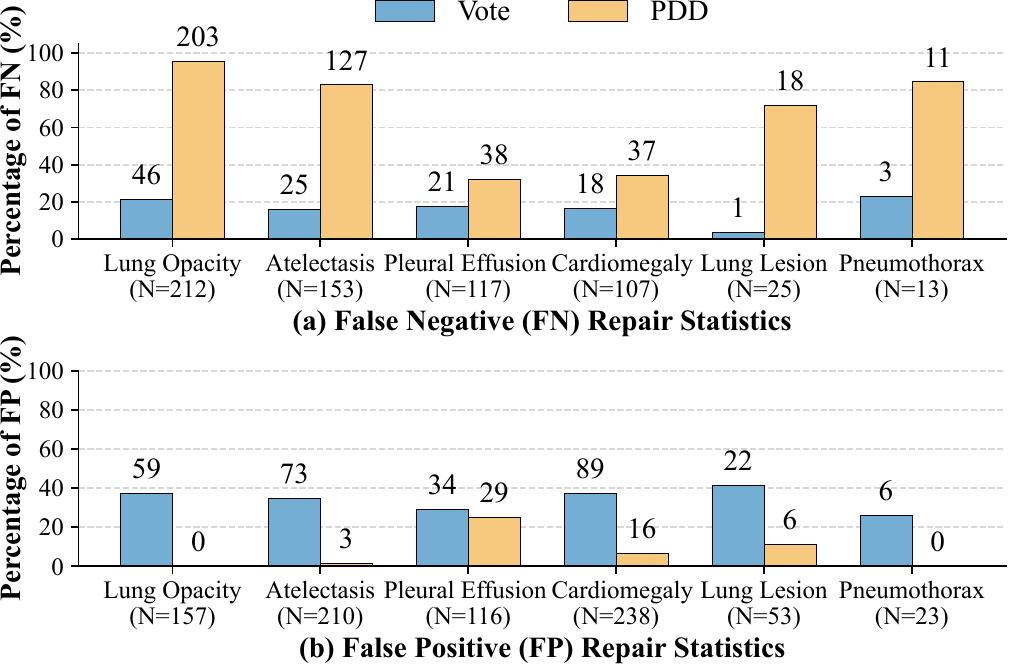}
    \caption{Error repair statistics for FN (a) and FP (b). The numbers above bars indicate the exact count of repaired samples.}
    \label{fig:Motivation-level_evidence}
\end{figure}

To investigate specific repair behaviors across diverse pathologies, we analyze error repair on the lesion-level. Due to space limits, Figure \ref{fig:Motivation-level_evidence} reports six representative diseases, with full results in our code repository.

Figure \ref{fig:Motivation-level_evidence}(b) shows that Voting is highly effective at FP suppression for both global findings such as Cardiomegaly and regional findings like Lung Opacity and Atelectasis. For global abnormalities, Voting leverages their expected cross-view consistency to eliminate projection artifacts; for regional findings, it acts as a denoiser, filtering out spurious single-view signals that lack corroboration.

However, a joint inspection of Figure \ref{fig:Motivation-level_evidence}(a) and (b) reveals a critical asymmetry: the categories where Voting excels at FP suppression are precisely those where PDD achieves massive FN recovery. This suggests that \emph{Voting’s “success” largely stems from discarding weak but genuine positive evidence}. Consequently, Voting fails to repair FNs effectively (e.g., Lung Opacity: 46 vs. 203 recovered by PDD).

This consensus bias \emph{is particularly harmful for focal and subtle lesions, where cross-view agreement is rare}. For Lung Lesion, Voting’s FN recovery collapses to near zero (1 case fixed), whereas PDD rescues 18 cases, \emph{an order of magnitude improvement}. 
Most critically, for Pneumothorax, a life-threatening emergency, PDD rescues 11 missed cases compared to only 3 by Voting, demonstrating its superior sensitivity to clinically critical abnormalities.

Overall, these lesion-level analyses reveal a fundamental trade-off: \emph{Voting favors conservatism by suppressing positives, while PDD-RRG prioritizes patient safety by recovering clinically meaningful findings}, especially when diagnostic evidence is sparse, localized, or view-dependent.

\section{Conclusion}
Our work provides the first systematic investigation of the post-processing decision layer for RRG. To address limitations in multi-view report generation, we propose the PDD-RRG, a likelihood-based aggregation framework that reconciles divergent inference paths from multiple input configurations of the same study. PDD-RRG enhances the use of clinically meaningful signals and unlocks the latent diagnostic capabilities of backbone models. Crucially, PDD-RRG requires no model retraining, effectively mitigates the information fusion bottlenecks in existing multi-view RRG systems. Nevertheless, PDD-RRG is inherently limited to resolving conflicts within the model’s own outputs. Consequently, it cannot fully correct severe hallucinations or intrinsic diagnostic blind spots. 
Future work will explore incorporating reliable external knowledge or multi-agent collaboration to overcome these limitations.




\section*{Acknowledgments}
This work was supported by Jiangsu Key Technology Research Development Program (BF2025036), and Hong Kong RGC grant GRF \#15611021. They are also with Jiangsu Key Lab of Language Computing, Suzhou.

\section*{Contribution Statement}
Yang Yu and Yiming Ji contributed equally to this work.


\bibliographystyle{named}
\bibliography{ijcai26}

\end{document}